\documentclass{article}
\usepackage{spconf,amsmath,graphicx}
\usepackage{textcomp}
\usepackage{xcolor}
\usepackage{multirow}
\usepackage{caption}
\usepackage{balance}
\usepackage{array}
\usepackage{graphicx}
\usepackage{subcaption}
\usepackage{pgfplots}
\usepackage{pgfplotstable}
\usepackage{sfmath}
\usepackage{lipsum}
\usepackage{graphicx}
\usepackage{color, colortbl}
\usepackage{graphicx}
\usepackage{tikz}
\usepackage{collcell}
\usepackage{etoolbox}
\usepackage{booktabs}
\usepackage{hhline}
\usepackage{pgf}
\usepackage{multirow}
\usepackage{graphicx}
\usepackage{caption}
\usepackage{subcaption}
\usepackage{tikz}

\title{Facial Action Unit Detection using 3D Facial Landmarks}
\name{Saurabh Hinduja and Shaun Canavan}
\address{University of South Florida}
\begin{document}
%
\maketitle
\begin{abstract}
In this paper, we propose to detect facial action units (AU)  using 3D facial landmarks. Specifically, we train a 2D convolutional neural network (CNN) on 3D facial landmarks, tracked using a shape index-based statistical shape model, for binary and multi-class AU detection. We show that the proposed approach is able to accurately model AU occurrences, as the movement of the facial landmarks corresponds directly to the movement of the AUs. By training a CNN on 3D landmarks, we can achieve accurate AU detection on two state-of-the-art emotion datasets, namely BP4D and BP4D+. Using the proposed method, we detect multiple AUs on over 330,000 frames, reporting improved results over state-of-the-art methods. 
\end{abstract}

\section{Introduction}
\label{sec:intro}
Facial geometry has a lot of information about an individual and has been used for various applications \cite{jeng1998facial}, \cite{kotsia2007facial}. It can also convey expressions, such as happy, sad, pain, and embarrassment \cite{zhang2016multimodal}, which can vary between subject to subject. Considering this, the Facial Action Coding System \cite{FACS} was developed, which represents fundamental facial activity in terms of Action Units (AUs). During an expression (i.e. facial activity), a single AU can occur or multiple at the same time. This allows for FACS to represent the large variety of facial expressions that exist between subjects.

Recently, there has been encouraging progress in automatically detecting AUs. Zeng et al. \cite{Zeng_2015_ICCV} developed a confidence preserving machine (CPM) for the task. In their proposed method, the CPM learns two classifiers. First the positive classifier separates all positive classes, and the negative classifier does the same for the negative classes. The CPM then learns a person-specific classifier to detect the AUs. Chu et al. \cite{chu2017learning} used a combination of convolutional neural networks (CNN) with long short-term networks to learn the spatial and temporal cues from images. Their proposed approach achieved promising results on the GFT \cite{cohn2010spontaneous} and BP4D \cite{zhang2014bp4d} datasets. Li et al. \cite{li2017action} used temporal fusing for AU detection. They developed a deep learning framework where regions of interest are learned independently so each sub-region has a local CNN; multi-label learning is then used.

 \begin{figure}[t]
 \label{fig:overview}
  \includegraphics[width=8.5cm, height=6cm]{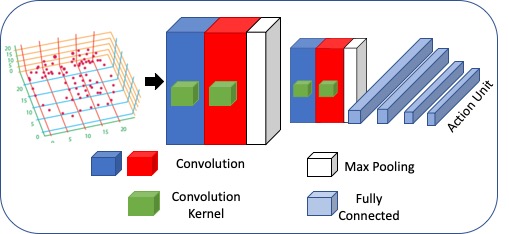}
  \caption{Proposed convolutional neural network pipeline for detecting action units from 3D facial landmarks.}
  \label{fig:overview}
\end{figure}

Current work, for detecting AUs, largely focuses on 2D images, however, it has been shown that 3D facial landmarks are important when understanding facial geometry for recognizing emotion \cite{fabiano2018}. Motivated by this, we propose to detect AUs using 3D facial landmarks. The contributions of this work are 3-fold and can be summarized as follows.
\begin{enumerate}
    \item 3D facial landmarks are used to detect action unit occurrences. Binary and 3-class classification experiments, with 3-fold and 10-fold cross-validation, are conducted to validate the proposed approach.
    \item The proposed approach of using 3D facial landmarks, outperforms current state-of-the-art, 2D image-based approaches, on the BP4D dataset \cite{zhang2014bp4d}.
    \item To test the generalizability of the proposed approach, cross dataset experiments are conducted, for AU detection, on the BP4D \cite{zhang2014bp4d}, and BP4D+ \cite{zhang2016multimodal} datasets.
\end{enumerate}

\section{Proposed Approach and Experimental Design}
\label{sec:method}
We propose detecting action units using 3D facial landmarks to train a convolutional neural network (CNN) for binary and 3-class classification. 
\subsection{Preprocessing 3D Facial Landmarks}
Given a set of 3D facial landmarks, \emph{L}, of size \emph{N}, where
\begin{equation}
L=(X_{1},Y_{1},Z_{1}),(X_{2},Y_{2},Z_{2}),....,(X_{N},Y_{N},Z_{N}),
\end{equation}
we first represent \emph{L} as a 2D matrix of size $N \times3$. We then normalize the 2D matrix of landmarks to be within the range between [0,1] using \emph{min-max} normalization. This is done, independently, for each of the \emph{(X,Y,Z)} axes of the frame as
\emph{$X_{norm}=\frac{(X_i-X_{min})}{(X_{max}-X_{min}}$}; \emph{$Y_{norm}=\frac{(Y_i-Y_{min})}{(Y_{max}-Y_{min}}$}; \emph{$Z_{norm}=\frac{(Z_i-Z_{min})}{(Z_{max}-Z_{min}}$}. 
These normalized values of $X_{norm}$, $Y_{norm}$, $Z_{norm}$ are then multiplied by a constant value \emph{C-1} to scale the landmarks into the range of \emph{[0,C-1]}, giving us $X_{scaled}, Y_{scaled}, Z_{scaled}$. This is done to scale all faces into the uniform range of \emph{[0,C-1]}. 

\begin{figure}
     \centering
     \begin{subfigure}[b]{4 cm}
         \centering
         \includegraphics[width=4cm]{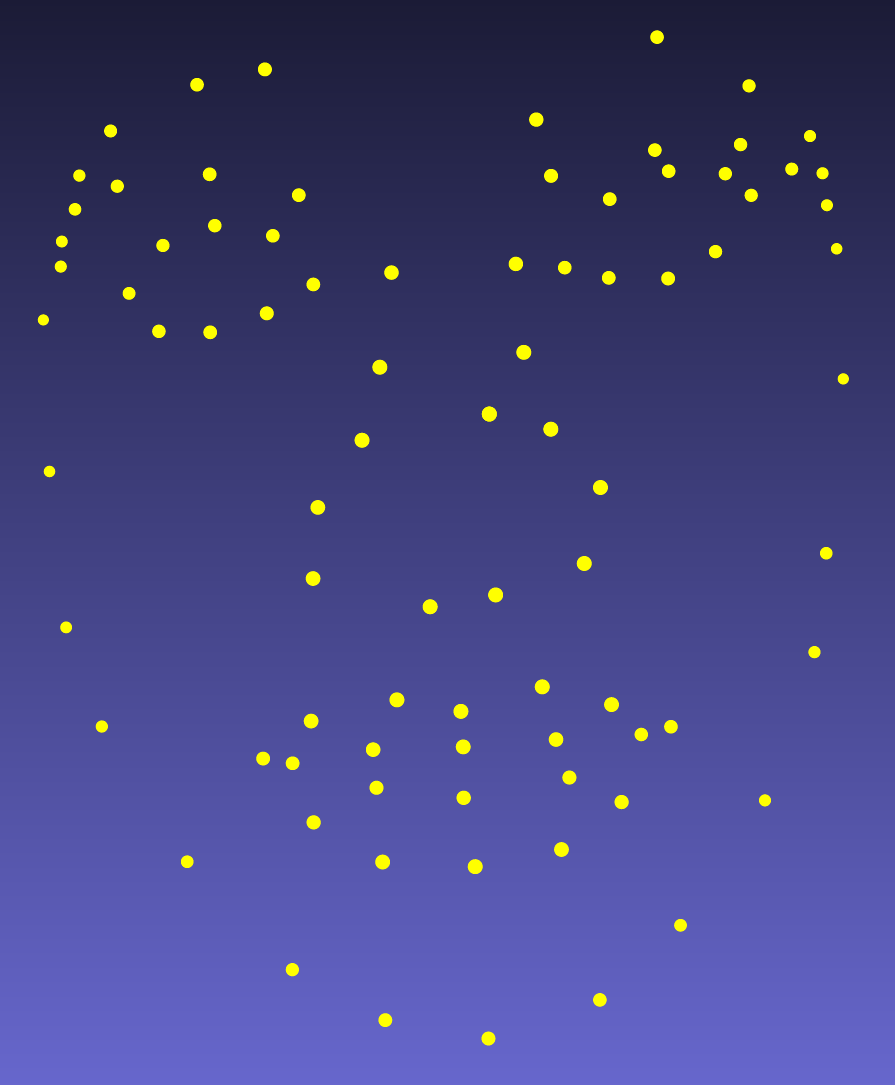}
         \caption{Original 3D Landmarks.}
         \label{fig:OriginaFace}
     \end{subfigure}
     \hfill
          \begin{subfigure}[b]{4 cm}
         \centering
         \includegraphics[width=4cm]{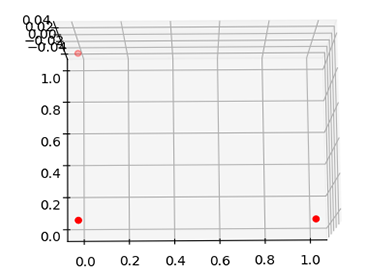}
         \caption{Non-scaled, normalized 3D mapping.}
         \label{fig:OriginaFace}
     \end{subfigure}
     \hfill
     \centering
     \begin{subfigure}[b]{4 cm}
         \centering
         \includegraphics[width=4cm]{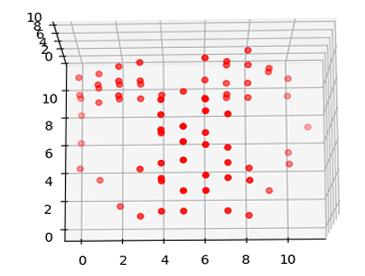}
         \caption{Scaled, Normalized 3D mapping with C=12.}
         \label{fig:OriginaFace}
     \end{subfigure}
     \hfill
     \begin{subfigure}[b]{4 cm}
         \centering
         \includegraphics[width=4 cm]{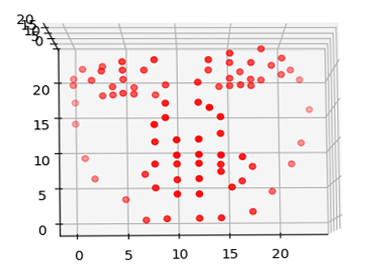}
         \caption{Scaled normalized 3D mapping with C=24.}
         \label{fig:3DFace}
     \end{subfigure}
        \caption{Comparison of original 3D landmarks with (a)non-scaled normalized 3D mapping (b) and with scaled normalized 3D Mapping for both C=12 (c) and C=24 (d).}
        \label{fig:FaceComparison}
\end{figure}

For the CNN to learn features for AU detection, we need $X_{scaled}, Y_{scaled}, Z_{scaled}$ to be scaled properly. Considering this we used a value of $C=24$ as we have empirically found that this gives a good representation of the face (Figure \ref{fig:FaceComparison} (d)). If we try to map the normalized (i.e. non-scaled) coordinates directly to a 3D array then the array will be of size $2 \times 2 \times 2$ each axis going from $[0,1]$. For example, if two or more normalized landmarks $(X_{norm}, Y_{norm}, Z_{norm})$ are in the range $[0,0.5)$ then they all set the same cell of $(0,0,0)$ to $1$, which can lead to loss in information of landmarks (Figure \ref{fig:FaceComparison} (b)). Also, if \emph{C} is not large enough, it can lead to loss of information (Figure \ref{fig:FaceComparison} (c)).

Given a set of scaled 3D facial landmarks, we then create a 3D representation of them, which can be used to train CNNs. To achieve this we create a 3D array \emph{A} of size $C \times C \times C$ initialized with all zeroes. We then set the locations where the landmarks are present to 1; $A[X_{scaled_i}][Y_{scaled_i}][Z_{scaled_i}]=1$. Scaling and mapping the landmarks creates a 3D array which is representative of the 3D locations of the landmarks (Figure \ref{fig:FaceComparison} (d)), which allows us to explicitly model the 3D shape of the landmarks, and subsequent AUs. As can be seen in Figure \ref{fig:FaceComparison}, the normalized 3D mapping does have some visual differences compared to the original 3D landmarks, however, the general shape, and more importantly AU activation, of the original 3D landmarks is the same. As we will show, this 3D representation of each set of 3D landmarks (i.e. face) can train CNNs to detect changes to the AUs across subjects, achieving high detection accuracy.

\subsection{Facial expression databases}
\label{sec:DB}
\textbf{BP4D} was used in the Facial Expression Recognition and Analysis (FERA) challenge in 2015  \cite{valstarFERA15} and 2017 \cite{valstarFERA17}; containing 2D and 3D data. It contains 41 subjects with eight dynamic expressions plus neutral. The dataset contains 18 male and 23 female subjects ages 18-29 years of age, with a range of ethnicities. For each sequence in this database, approximately 500 frames contain AU occurrences; we used all labeled frames, which is over 140,000.\\
\textbf{BP4D+} consists of 140 subjects (58 males and 82 females); ages 18-66, each with highly varied emotional responses. It includes thermal, physiological, 2D, and 3D data. It was also used in the FERA challenge 2017 \cite{valstarFERA17}. Like BP4D, approximately 500 frames per sequence contain AU occurrences; we used all labeled frames which is over 190,000.

\subsection{Experimental Design}
\label{sec:expDesign}
We detected 83 3D facial landmarks on all AU labeled frames from BP4D and BP4D+, using a shape-index-based statistical shape model \cite{canavan2015landmark}. We then normalized the 83 landmarks to [0,23] ($C=24$). While any number of landmarks that represent a face with AU occurrences can be used, we chose 83 to be consistent with other related works \cite{fabiano2018, valstarFERA17, zhang2014bp4d, zhang2016multimodal}. We then transform the normalized landmarks to a binary representation in a 3 dimensional array of $24 \times 24 \times 24$ (Figure \ref{fig:overview} (d)).

We consider both binary \cite{li2017action} and 3-class \cite{chu2017learning} classification. For binary classification, the presence/absence of an action unit was coded as 1/0. For the 3-class classification, the presence/absence of an action unit was coded as 1/-1 and 0 when no information was available \cite{chu2017learning}. 

For the binary-class, our CNN consists of 8 layers; two convolutional layers followed by a max pooling layer, two more convolutional layers, another max pooling layer and finally three fully connected layers. The output layer has 12 binary outputs for the 12 AUs being predicted. The activation function for each of the hidden layers is relu \cite{jarrett2009best} and the activation function for the output layer is sigmoid \cite{han1995influence}. T   he loss function for this network was binary cross-entropy and the optimizer used was adam \cite{kingma2014adam}, where all training is done with 250 epochs. See Figure \ref{fig:overview} for proposed pipleline and network architecture

For the 3-class, we use a similar network with common convolutional
layers, but different fully connected layers for each AU (a common output layer can not predict the 3-classes for all the AUs). The loss function was categorical cross-entropy and the activation function for the initial layers was still relu, but for the output layers it was softmax \cite{bridle1990probabilistic}, as it  performs well for multi-class classification \cite{duan2003multi}. Again, all training is done with 250 epochs.

For binary-class, we used 3-fold and 10-fold cross-validation, for 3-class problem we used 3-fold cross-validation. We also balanced the distribution of positive and negative samples (i.e. occurrence of AUs), as the distribution of AUs is not consistent (Table \ref{Distribution}). We chose this experimental design to be consistent with related works \cite{chu2017learning}, \cite{li2017action}. As can be seen from Table \ref{Distribution}, some of the AUs have a small number of frames where the AU occurred, especially in BP4D+.
\begin{table}
\centering
\captionsetup{justification=centering}
 \begin{tabular}{|p{0.5cm}|p{1.2 cm}|p{1.2 cm}|}
\hline 
AU &BP4D &BP4D+ \\ \hline 
1	&21.07	&9.54\\ \hline 
2	&17.04	&8.01\\ \hline 
4	&20.22	&0.67\\ \hline 
6	&46.10	&65.88\\ \hline 
7	&54.90	&3.46\\ \hline 
10	&59.39	&57.37\\ \hline 
12	&56.18	&59.99\\ \hline 
14	&46.60	&32.46\\ \hline 
15	&16.96	&12.96\\ \hline 
17	&34.37	&0.67\\ \hline 
23	&16.56	&2.35\\ \hline 
24	&15.16	&-\\ \hline
\end{tabular}
\centering
\caption{Percentage of AU labeled frames with occurrences in BP4D and BP4D+.}
\label{Distribution}
\end{table}

For binary classification, we conducted 4 experiments for each 3-fold and 10-fold: (1) training and testing on BP4D; (2) training and testing on BP4D+;  (3) training on BP4D and testing on BP4D+; and (4) training on BP4D+ and testing on BP4D. For our 3-class problem, we performed 2 experiments: (1) training and testing on BP4D; and (2) training and testing on BP4D+. For BP4D we detected 12 AUs, and 11 for BP4D+, as AU 24 did not occur in the labeled frames of this dataset. We refer the reader to Tables \ref{table:BinaryF1}, \ref{table:Cross Validation}, or \ref{table:3ClassResults} for the list of AUs detected for each dataset.

With this type of classification, especially with imbalanced data (Table \ref{Distribution}), F1 score can be a better indicator of performance compared to classification accuracy \cite{valstarFERA15}. Considering this, we calculated the frame-based F1 score as ($F1-frame=\frac{2RP}{R+P}$) where R is recall and P is precision. This approach is also consistent with related works, allowing us to compare our results.
\section{Results and Analysis}
\label{sec:results}

\subsection{Binary classification}
\label{sec:binClass}
For binary classification on BP4D, we achieved an average F1 binary score of 92.90 and 94.08 for 3-fold and 10-fold, respectively. On BP4D+, we achieved an average F1 binary score of 86.01 and 87.63 for 3-fold and 10-fold, respectively. See Table \ref{table:BinaryF1} for more details.
\begin{table}
\centering
\begin{tabular}{ |p{0.7cm}||p{1.3cm}|p{1.3cm}||p{1.3cm}|p{1.3cm}|  }
 \hline 
 \multirow{2}{*}{AU} & \multicolumn{2}{c||}{BP4D} & \multicolumn{2}{c|}{BP4D+}\\ \cline{2-5}&3 Fold &10 Fold &3 Fold &10 Fold  \\ \hline
1	&91.16	&92.68	&82.78 &85.24\\ \hline 
2	&90.31	&92.11	&82.82	&85.21\\ \hline
4	&93.12	&94.14	&75.57	&78.07\\ \hline
6	&96.24	&97.01	&97.18	&97.50\\ \hline
7	&96.40	&96.98	&80.62	&83.67\\ \hline
10	&97.59	&97.95	&96.97	&97.35\\ \hline
12	&97.89	&98.31	&95.85	&96.40\\ \hline
14	&95.47	&96.29	&91.22	&92.31\\ \hline
15	&87.63	&89.66	&83.26	&85.19\\ \hline
17	&91.14	&92.35	&72.39	&73.62\\ \hline
23	&85.87	&88.23	&87.40	&89.39\\ \hline
24	&91.93	&93.19	&-	&-  \\ \hline
\textbf{Avg}	&\textbf{92.90}	&\textbf{94.08}	&\textbf{86.01}	&\textbf{87.63}\\ \hline 
\end{tabular}
\captionsetup{justification=centering}
\caption{Binary F1 scores for 3-fold and 10-fold on BP4D and BP4D+.}
\label{table:BinaryF1}
\end{table}

 We also investigated cross-database validation between BP4D and BP4D+. When training on BP4D and testing on BP4D+, we achieved an average F1 score of 42.9 and 42.84 for 3-fold and 10-fold, respectively. When training on BP4D+ and testing on BP4D, we achieved an average F1 score of 39.1 and 40.02 for 3-fold and 10-fold, respectively (Table \ref{table:Cross Validation}). We found that the best performing AUs were 6, 10, and 12; while the worst performing were 4, 17, and 23. This difference in F1 scores can be explained by the disparity in the occurrence of AUs. The best performing AUs are present in a similar percent of the frames whereas the poor performing AUs have a large difference between BP4D and BP4D+ (Table \ref{Distribution}). The biggest difference in occurrence is for AU 17; BP4D has 34.37\% and BP4D+ has just 0.67\% frames with the AU. 
 
 \begin{table}
\centering
\captionsetup{justification=centering}
\newcolumntype{C}[1]{>{\centering\arraybackslash}p{#1}}
\begin{tabular} { |p{0.7cm}||p{1.3cm}|p{1.3cm}||p{1.3cm}|p{1.3cm}|  }
\hline
\multirow{2}{*}{AU} & \multicolumn{2}{C{2.6cm}||}{Trained BP4D+ / Tested BP4D} & \multicolumn{2}{C{2.6cm}|}{Trained BP4D/ Tested BP4D+}\\ \cline{2-5}&3 Fold &10 Fold &3 Fold &10 Fold  \\ \hline
1	&47.30	&54.27	&54.28	&53.18 \\ \hline
2	&43.50	&47.11	&44.05	&42.53 \\ \hline
4	&12.99	&11.44	&11.83	&13.06 \\ \hline
6	&70.05	&69.79	&79.24	&79.05 \\ \hline
7	&23.85	&20.72	&26.73	&26.74 \\ \hline
10	&72.54	&73.33	&80.46	&80.41 \\ \hline
12	&75.61	&75.21	&76.60	&76.77 \\ \hline
14	&36.49	&36.30	&41.54	&41.81 \\ \hline
15	&23.55	&24.30	&22.53	&22.72 \\ \hline
17	&10.03	&12.32	&14.61	&14.26 \\ \hline
23	&14.26	&15.38	&20.06	&20.68 \\ \hline
\textbf{Avg}	&\textbf{39.10}	&\textbf{40.02}	&\textbf{42.90}	&\textbf{42.84} \\ \hline
\end{tabular}
\centering
\caption{Cross-database F1 scores for 3-fold and 10-fold.}
\label{table:Cross Validation}
\end{table}
 
\subsection{3-class classification}
We performed 3-fold validation on BP4D and BP4D+, reporting the F1-macro and micro scores (Table \ref{table:3ClassResults}). F1-macro is the average of the F1 scores of the 3 classes; where as F1-micro is the weighted average of the 3 F1 scores. On BP4D, we achieved an average F1-micro score of 94.55, and F1-macro score of 96.09. For BP4D+, an F1-micro score of 87.90 and F1-macro score of 97.03 was achieved. Some of the lowest performing AUs with BP4D+ were again 4, 17, and 23. This is consistent with the cross-database validation and can be explained by the low number of AU occurrences (Table \ref{Distribution}).
\begin{table}
\centering
\begin{tabular}{ |p{0.6cm}||p{1.4cm}|p{1.4cm}||p{1.4cm}|p{1.4cm}|  }
 \hline 
 \multirow{2}{*}{AU} & \multicolumn{2}{c||}{BP4D} & \multicolumn{2}{c|}{BP4D+}\\ \cline{2-5}&F1 Macro &F1 Mirco &F1 Macro &F1 Mirco  \\ \hline
 
1	&93.81	&96.01	&86.90	&96.57\\ \hline
2	&93.67	&96.52	&87.06	&97.07\\ \hline
4	&95.12	&96.93	&85.83	&99.61\\ \hline
6	&96.18	&96.21	&91.69	&96.02\\ \hline
7	&95.66	&95.71	&86.62	&98.60\\ \hline
10	&96.79	&96.90	&90.51	&96.22\\ \hline
12	&97.33	&97.37	&90.80	&94.66\\ \hline
14	&95.60	&95.62	&89.17	&94.15\\ \hline
15	&91.88	&95.65	&87.39	&95.54\\ \hline
17	&92.80	&93.58	&82.19	&99.54\\ \hline
23	&90.79	&95.17	&88.71	&99.30\\ \hline
24	&94.94	&97.46	&-	&-\\ \hline
\textbf{Avg}	&\textbf{94.55}	&\textbf{96.09}	&\textbf{87.90}	&\textbf{97.03}\\ \hline
    \end{tabular}
    \caption{F1 scores for 3-class, 3-fold cross validation.}
    \label{table:3ClassResults}
\end{table}
\subsection{Comparisons to state of the art}
On BP4D, many works use 2D images for 3-fold binary \cite{li2017action}, 10-fold binary \cite{Zeng_2015_ICCV}, and 3-fold 3-class classification \cite{chu2017learning}. For each of these experimental designs, the proposed method outperforms the state of the art, detailing the power of explicitly using 3D facial landmarks compared to 2D images. 

For 3-fold binary classification, the proposed method achieves a significant increase in the average F1 score compared to current state of the art (Table \ref{table:Comparison}). For 10-fold binary classification, the proposed method achieved an average F1 score of 94.08, compared to Zeng et al. \cite{Zeng_2015_ICCV}, that achieved an average F1 score of 56.5. We also compare our 3-fold, 3-class results to state of the art. The proposed method achieved an average F1-micro and macro score of 96.06 and 94.55, respectively, compared to the work from Chu et al. \cite{chu2017learning} that reported an average F1 score of 82.5. These increases can be attributed to the proposed 3D representation of the landmarks.
\begin{table}
\centering
\begin{tabular}{|p{2.5cm}|p{2cm}|}
\hline
Method &Avg F1 Score\\ \hline
\textbf{Proposed}	&\textbf{92.9}\\ \hline
R-T1\cite{li2017action}	&66.1\\ \hline
FERA\cite{jaiswal2016deep}	&61.4\\ \hline
CNN+LSTM\cite{chu2016modeling}	&53.2\\ \hline
CPM\cite{Zeng_2015_ICCV}	&50.0\\ \hline
DRML\cite{Zhao_2016_CVPR}	&48.3\\ \hline
JPML\cite{Zhao_2015_CVPR}	&45.9\\ \hline
\end{tabular}
\caption{Comparison of proposed method with state of the art on BP4D, for 3-fold binary classification.}
\label{table:Comparison}
\end{table}
For BP4D+, a subset of data was used in FERA 2017 \cite{valstarFERA17}. BP4D was used as training data, and BP4D+ was used as development and testing sets. They report results, using maximum likelihood, on both of these sets. When training on BP4D and testing on BP4D+, the proposed method achieved an average F1 score of 42.9 and 42.84, for 3-fold and 10-fold cros-validation, respectively. This compares to an average F1 score of 41.6 and 45.2 on the FERA development and test sets, respectively. As the two experimental design are \textit{not} the same, we \textit{do not} claim this as a direct comparison. It is included for clarification of results reported on BP4D+.

\section{Conclusion}
Detecting action units is an important task in face analysis, especially in facial expression recognition. This is due, in part, to the idea that expressions can be decomposed into multiple action units. Considering this, we have proposed detecting action units using 3D facial landmarks to train a convolutional neural network. Experimental results on binary and 3-class classification show encouraging AU detection results on the BP4D and BP4D+ datasets. We have also conducted cross-database validation, of the proposed approach, by training on BP4D and testing on BP4D+, as well as training on BP4D+ and testing on BP4D. We report state of the art results on BP4D using 3-fold and 10-fold cross-validation.
\section*{Acknowledgment}
We gratefully acknowledge the support of NVIDIA Corporation with a donation of a Titan XP GPU used in this research.
\balance
\bibliographystyle{IEEE}
\bibliography{references}


\end{document}